\documentclass[11pt]{article}

\usepackage[final]{acl}
\usepackage{booktabs}
\usepackage{times}
\usepackage{latexsym}
\usepackage{makecell}
\usepackage{amsfonts}
\usepackage[T1]{fontenc}
\newcommand{\minimize}{\operatorname*{minimize}}

\newcommand{\argmin}{\mathop{\rm argmin}}

\usepackage[utf8]{inputenc}
\usepackage{algorithm}
\usepackage{algpseudocode}
\usepackage{microtype}
\usepackage{tikz}
\usetikzlibrary{calc}
\usepackage{amsmath}
\usepackage{multirow}
\usepackage{enumitem}

\usepackage{pgfplots}
\usepackage{pgfplotstable}
\pgfdeclarelayer{front}
\pgfdeclarelayer{back}
\pgfsetlayers{back,main,front}

\usetikzlibrary{fit,positioning,shapes,backgrounds,arrows.meta,calc}
\usepgfplotslibrary{groupplots,fillbetween,colorbrewer}
\usepackage{caption,subcaption}

\usepackage{inconsolata}
\usepackage{amsmath}

\usepackage{amssymb}
\usepackage{amsthm}
\usepackage{hyperref}
\usepackage[capitalize]{cleveref}

\usepackage{graphicx}
\theoremstyle{plain}
\newtheorem{theorem}{Theorem}[section]

\theoremstyle{definition}
\newtheorem{definition}[theorem]{Definition}
\crefname{definition}{Definition}{Definitions}
\theoremstyle{remark}
\newtheorem{remark}[theorem]{Remark}
\crefname{remark}{Remark}{Remarks}
\definecolor{bg}{RGB}{248,251,254}
\definecolor{frame}{RGB}{157,173,193}
\definecolor{line}{RGB}{105,112,118}
\definecolor{lightline}{RGB}{190,196,202}
\definecolor{chipbg}{RGB}{255,244,200}
\definecolor{taghi}{RGB}{255,224,145}
\definecolor{stripborderhi}{RGB}{205,155,30}
\definecolor{borderhi}{RGB}{220,70,60}
\definecolor{xhighlight}{RGB}{195,140,20}

\definecolor{qone}{RGB}{105,200,145}      % scheme 1: w1a16_g128_asym
\definecolor{qtwo}{RGB}{105,190,200}     % scheme 2: w2a16_g128_asym
\definecolor{qthree}{RGB}{105,117,200}   % scheme 3: w3a16_g128_asym
\definecolor{qfour}{RGB}{160,105,200}    % scheme 4: w4a16_g128_asym

\tikzset{
  font=\sffamily,
  every node/.style={text=black},
  outer/.style={
    rounded corners=8pt,
    draw=frame,
    fill=bg,
    line width=0.8pt
  },
  panel/.style={
    rounded corners=6pt,
    draw=line!80,
    fill=white,
    line width=0.6pt
  },
  bank/.style={
    rounded corners=7pt,
    draw=borderhi,
    fill=white,
    line width=1.5pt
  },
  row/.style={
    rounded corners=4pt,
    draw=line!65,
    fill=white,
    line width=0.55pt
  },
  constraint/.style={
    rounded corners=3pt,
    draw=line!65,
    dashed,
    fill=white,
    line width=0.50pt
  },
  tag/.style={
    rounded corners=3pt,
    draw=frame,
    fill=white,
    line width=0.55pt
  },
  legendgroup/.style={
    rounded corners=5pt,
    draw=line!75,
    fill=white,
    line width=0.6pt
  },
  legendbox/.style={
    draw=black,
    line width=1.0pt
  },
  thumbhi/.style={
    rounded corners=4pt,
    draw=borderhi,
    fill=white,
    line width=1.5pt
  },
  stripbox/.style={
    rounded corners=3pt,
    draw=line!65,
    line width=0.45pt
  },
  arrow/.style={
    -{Latex[length=2.2mm,width=1.7mm]},
    draw=line!85,
    line width=0.65pt
  }
}

\newcommand{\SetSchemeColor}[1]{%
  \ifcase#1\relax
  \or\def\CurrentSchemeColor{qone}%
  \or\def\CurrentSchemeColor{qtwo}%
  \or\def\CurrentSchemeColor{qthree}%
  \or\def\CurrentSchemeColor{qfour}%
  \fi
}

\newcommand{\LegendItem}[4]{%
  \draw[legendbox,fill=#3] (#1,#2) rectangle ++(0.30,0.30);
  \node[anchor=west,font=\fontsize{10}{12}\selectfont] at
    ($(#1,#2)+(0.40,0.15)$) {#4};
}

\newcommand{\LayerThumb}[7]{%
  \pgfmathsetmacro{\thumbTop}{#2 + 2.65}
  \pgfmathsetmacro{\thumbBot}{#2 - 2.03}
  \draw[#5] (#1 - 2.30,\thumbBot) rectangle (#1 + 2.30,\thumbTop);
  \node[font=\bfseries\fontsize{14}{16}\selectfont]
    at (#1,\thumbTop - 0.32) {#7};
  \foreach \sIdx/\bw/\sYoff/\ov in {0/1/0.65/1, 1/2/2.05/0, 2/3/3.45/2} {
    \pgfmathsetmacro{\sTop}{\thumbTop - \sYoff}
    \ifnum\sIdx=#6\relax
      \draw[stripbox,fill=chipbg,draw=stripborderhi,line width=1.5pt]
        (#1 - 2.20,\sTop - 1.18) rectangle (#1 + 2.20,\sTop + 0.08);
    \else
      \draw[stripbox]
        (#1 - 2.20,\sTop - 1.18) rectangle (#1 + 2.20,\sTop + 0.08);
    \fi
    \ifnum\sIdx=#6\relax
      \draw[tag,fill=taghi,draw=stripborderhi,line width=1.5pt]
        (#1 - 2.10,\sTop - 0.87) rectangle (#1 - 0.65,\sTop - 0.27);
    \else
      \draw[tag]
        (#1 - 2.10,\sTop - 0.87) rectangle (#1 - 0.65,\sTop - 0.27);
    \fi
    \node[font=\large]
      at (#1 - 1.375,\sTop - 0.57) {$\mathbf{x}_{#4}^{\star}(\beta^{(\bw)})$};
    \node[font=\fontsize{8}{9}\selectfont]
      at (#1 - 0.50,\sTop - 0.15) {u};
    \node[font=\fontsize{8}{9}\selectfont]
      at (#1 - 0.50,\sTop - 0.45) {g};
    \node[font=\fontsize{8}{9}\selectfont]
      at (#1 - 0.50,\sTop - 0.75) {d};
    \foreach \vis/\colNum in {0/1, 1/2, 2/3, 3/4, 4/5, 5/6, 6/7, 7/8} {
      \foreach \r in {0,1,2} {
        \pgfmathtruncatemacro{\base}{mod(\colNum + 2*\r + #3 + \sIdx + floor((\colNum + #3 + \sIdx)/7),100)}
        \pgfmathtruncatemacro{\sch}{%
          (\ov == 0) ? (mod(\base,4) + 1) :
          ((\ov == 1) ? ((mod(\base,10) < 7) ? (mod(\base,2) + 1) : (mod(\base,2) + 3)) :
            ((\ov == 2) ? ((mod(\base,10) < 7) ? ((mod(\base,3) < 1) ? 3 : 4) : (mod(\base,2) + 1)) :
              \ov))}
        \SetSchemeColor{\sch}
        \pgfmathsetmacro{\cX}{#1 - 0.35 + \vis*0.30}
        \pgfmathsetmacro{\cY}{\sTop - \r*0.30}
        \draw[fill=\CurrentSchemeColor,draw=black,line width=1.0pt]
          (\cX,\cY - 0.30) rectangle ++(0.30,0.30);
      }
    }
    \foreach \vis/\col in {0/1, 1/2, 2/3, 3/4, 4/5, 5/6, 6/7, 7/8} {
      \pgfmathsetmacro{\indX}{#1 - 0.20 + \vis*0.30}
      \node[font=\fontsize{8}{9}\selectfont]
        at (\indX,\sTop - 1.04) {\col};
    }
  }
}

\newcommand{\LowerLayerOverview}{%
  \draw[outer] (0.20,-5.33) rectangle (20.99,0.40);
  \node[font=\bfseries\fontsize{22}{24}\selectfont,fill=bg,inner sep=3pt]
    at (10.595,0.00)
    {Per-block candidate caches $\{\mathcal{S}_l\}_{l=1}^{L}$};
  \DrawSchemeLegend
  \LayerThumb{6.50}{-3.10}{4}{1}{row}{0}{{\fontsize{12}{14}\selectfont Block $1$ assignment }$\mathcal{S}_1$}
  \node[font=\Large] at (9.45,-3.10) {$\cdots$};
  \LayerThumb{12.40}{-3.10}{0}{l}{thumbhi}{2}{{\fontsize{12}{14}\selectfont Block $l$ assignment }$\mathcal{S}_l$}
  \draw[fill=qone,draw=black,line width=1.0pt] (14.15,-1.40) rectangle (14.45,-1.10);
  \draw[fill=qone,draw=black,line width=1.0pt] (13.25,-1.70) rectangle (13.55,-1.40);
  \draw[fill=qone,draw=black,line width=1.0pt] (12.65,-2.00) rectangle (12.95,-1.70);
  \node[font=\Large] at (15.35,-3.10) {$\cdots$};
  \LayerThumb{18.30}{-3.10}{16}{L}{row}{1}{{\fontsize{12}{14}\selectfont Block $L$ assignment }$\mathcal{S}_L$}
}

\newcommand{\DrawSchemeLegend}{%
  \draw[legendgroup] (0.40,-5.13) rectangle (4.00,-0.45);
  \node[font=\bfseries\fontsize{14}{16}\selectfont]
    at (2.20,-1.00) {Quantizer set $\mathcal{Q}$};
  \LegendItem{0.90}{-1.90}{qone}{w1a16\_g128\_asym}
  \LegendItem{0.90}{-2.70}{qtwo}{w2a16\_g128\_asym}
  \LegendItem{0.90}{-3.50}{qthree}{w3a16\_g128\_asym}
  \LegendItem{0.90}{-4.30}{qfour}{w4a16\_g128\_asym}
}

\newcommand{\LayerSolutionBankFinal}{%
  \path[use as bounding box] (0,-5.43) rectangle (21.20,7.60);
  \draw[outer] (0.20,0.50) rectangle (10.595,7.50);
  \node[font=\bfseries\fontsize{22}{24}\selectfont,fill=bg,inner sep=3pt]
    at (5.3975,7.10)
    {(1) Inner stage (per-block)};
  \node[align=center,font=\fontsize{18}{22}\selectfont] at (5.3975,6.20)
    {Collect Pareto-optimal per-block assignments $\mathcal{S}_l$ \\
     for each budget level $\beta^{(k)} \in \mathcal{B}$ using proxy.};
  \draw[panel] (0.40,0.70) rectangle (5.595,5.38);
  \draw[arrow] (1.10,1.88) -- (5.40,1.88);
  \draw[arrow] (1.10,1.88) -- (1.10,5.18);
  \foreach \tx/\lab in {2.15/{$\beta^{(1)}$}, 3.20/{$\beta^{(2)}$}, 4.25/{$\beta^{(3)}$}} {
    \draw[line!85,line width=0.6pt] (\tx,1.78) -- (\tx,1.88);
    \node[font=\footnotesize,anchor=north] at (\tx,1.78) {\lab};
  }
  \node[font=\large] at (3.20,1.18) {Bitwidth};
  \node[font=\large,rotate=90] at (0.70,3.53) {Proxy};
  \foreach \px/\py in {2.15/4.80, 2.15/4.30, 2.15/3.95,
                        3.20/4.55, 3.20/3.95, 3.20/3.40,
                        4.25/4.50, 4.25/3.50, 4.25/3.10} {
    \fill[red!40] (\px,\py) circle (0.06);
  }
  \draw[red,line width=0.8pt] (2.15,3.60) -- (3.20,2.80) -- (4.25,2.50);
  \foreach \px/\py in {2.15/3.60, 3.20/2.80, 4.25/2.50} {
    \fill[red] (\px,\py) circle (0.08);
  }
  \node[anchor=west,font=\Large] at (4.40,2.50) {$\mathbf{x}_l^{\star}$};
  \LayerThumb{8.095}{2.73}{0}{l}{thumbhi}{-1}{{\fontsize{12}{14}\selectfont Block $l$ assignment }$\mathcal{S}_l$}
  \draw[fill=qone,draw=black,line width=1.0pt] (9.845,4.43) rectangle (10.145,4.73);
  \draw[fill=qone,draw=black,line width=1.0pt] (8.945,4.13) rectangle (9.245,4.43);
  \draw[fill=qone,draw=black,line width=1.0pt] (8.345,3.83) rectangle (8.645,4.13);
  \draw[red,dashed,line width=0.7pt] (2.15,3.60) -- (5.895,4.18);
  \draw[red,dashed,line width=0.7pt] (3.20,2.80) -- (5.895,2.78);
  \draw[red,dashed,line width=0.7pt] (4.25,2.50) -- (5.895,1.38);
  \draw[outer] (10.795,0.50) rectangle (20.99,7.50);
  \node[font=\bfseries\fontsize{22}{24}\selectfont,fill=bg,inner sep=3pt]
    at (15.8925,7.10)
    {(2) Outer stage (global)};
  \node[align=center,font=\fontsize{18}{22}\selectfont] at (15.8925,6.20)
    {Solve how much budget $\beta_{l}$ to allocate on each \\
     MoE block $l$ using the model-level objective.};
  \node[font=\fontsize{16}{20}\selectfont] at (15.8925,4.20)
    {$\begin{aligned}\boldsymbol{\beta}^{\star} ={}& \mathop{\mathrm{argmin}}_{\boldsymbol{\beta} \in \mathcal{B}^L} \left( \mathcal{L}(\mathbf{x}(\boldsymbol{\beta})) := \mathcal{L}(\mathbf{x}_1^{\star}(\beta_{1}), \ldots, \mathbf{x}_L^{\star}(\beta_{L})) \right) \\[-2pt] & \hspace{0.2em} \text{subject to} \quad \frac{1}{L}\sum_{l=1}^{L} \beta_l \leq \tau \end{aligned}$};
  \node[font=\fontsize{16}{20}\selectfont] at (15.8925,2.20) {Get global assignment};
  \node[anchor=west,font=\fontsize{16}{20}\selectfont] at (10.95,1.50)
    {$\mathbf{x}^{\star} = \mathbf{x}^{\star}(\boldsymbol{\beta}^{\star}) =$};
  \node[anchor=west,font=\fontsize{16}{20}\selectfont] at (13.85,1.50) {$($};
  \draw[tag,fill=taghi,draw=stripborderhi,line width=1.5pt]
    (14.27,1.20) rectangle (15.72,1.80);
  \node[font=\large] at (14.995,1.50) {$\mathbf{x}_1^{\star}(\beta_1^{\star})$};
  \node[anchor=west,font=\fontsize{12}{14}\selectfont] at (15.77,1.50)
    {$,\,\ldots\,,$};
  \draw[tag,fill=taghi,draw=stripborderhi,line width=1.5pt]
    (16.72,1.20) rectangle (18.17,1.80);
  \node[font=\large] at (17.445,1.50) {$\mathbf{x}_l^{\star}(\beta_l^\star)$};
  \node[anchor=west,font=\fontsize{12}{14}\selectfont] at (18.22,1.50)
    {$,\,\ldots\,,$};
  \draw[tag,fill=taghi,draw=stripborderhi,line width=1.5pt]
    (19.17,1.20) rectangle (20.62,1.80);
  \node[font=\large] at (19.895,1.50) {$\mathbf{x}_L^{\star}(\beta_L^\star)$};
  \node[anchor=west,font=\fontsize{16}{20}\selectfont] at (20.65,1.50) {$)$};
  \LowerLayerOverview
  \draw[stripborderhi,dashed,line width=0.8pt] (8.70,-1.02) -- (14.27,1.20);
  \draw[stripborderhi,dashed,line width=0.8pt] (14.60,-3.82) -- (16.72,1.20);
  \draw[stripborderhi,dashed,line width=0.8pt] (20.50,-2.42) -- (20.62,1.20);
}

\title{Q-Strata: Hierarchical Bit Allocation for Mixed-Precision Quantization of Mixture-of-Experts LLMs}

\author{Deokjae Lee\textsuperscript{1,3} \and Sihun Chu\textsuperscript{2,3} \and Hyun Oh Song\textsuperscript{1,2,3\,*} \\
  \textsuperscript{1}Department of Computer Science and Engineering, Seoul National University \\
  \textsuperscript{2}Interdisciplinary Program in Artificial Intelligence, Seoul National University \\
  \textsuperscript{3}Neural Processing Research Center \\
  \texttt{\{bdbj, sihun.chu, hyunoh\}@mllab.snu.ac.kr}}

\begin{document}
\maketitle
\begingroup
\renewcommand{\thefootnote}{*}%
\footnotetext{Hyun Oh Song is the corresponding author.}%
\endgroup
\begin{abstract}
Mixed-precision quantization (MPQ) assigns a different bitwidth to each linear layer of a large language model (LLM) to minimize the quantization-induced quality loss under a fixed budget, but Mixture-of-Experts (MoE) models contain these layers in every expert of every MoE block, so the allocation space grows far larger than in a dense model.
Existing methods either allocate within each block under a uniform per-block budget, or allocate across blocks through an additive proxy, and neither directly optimizes a model-level objective over the choices that couple the blocks.
We propose \textsc{Q-Strata}, a bi-level allocator that ranks within-block assignments with a cheap proxy and allocates across blocks with a model-level objective evaluated on the assembled quantized model.
Its inner stage caches a Pareto frontier of candidates per block over finely spaced budgets, leaving the outer stage to set one budget per block instead of a bitwidth for every linear layer.
With the search reduced to one budget per block, the outer stage optimizes this model-level objective directly, capturing the inter-block coupling that additive proxies miss.
On Mixtral-8x7B-Instruct, Qwen1.5-MoE-A2.7B, and DeepSeek-V2-Lite, \textsc{Q-Strata} consistently achieves lower WikiText2 perplexity than uniform-bitwidth GPTQ and the state-of-the-art MoE MPQ methods MxMoE and GEMQ in the low-bit regime.
The code is available at \url{https://github.com/snu-mllab/Q-Strata/tree/main}.

\end{abstract}
\section{Introduction}
\label{sec:intro}

% \begin{figure*}[t]
%     \centering
%     % \includegraphics[width=\textwidth]{figures/your_figure.pdf}
%     \fbox{\rule{0pt}{7cm}\rule{\textwidth}{0pt}} % placeholder
%     \caption{Figure placeholder spanning both columns.}
%     \label{fig:wide}
% \end{figure*}

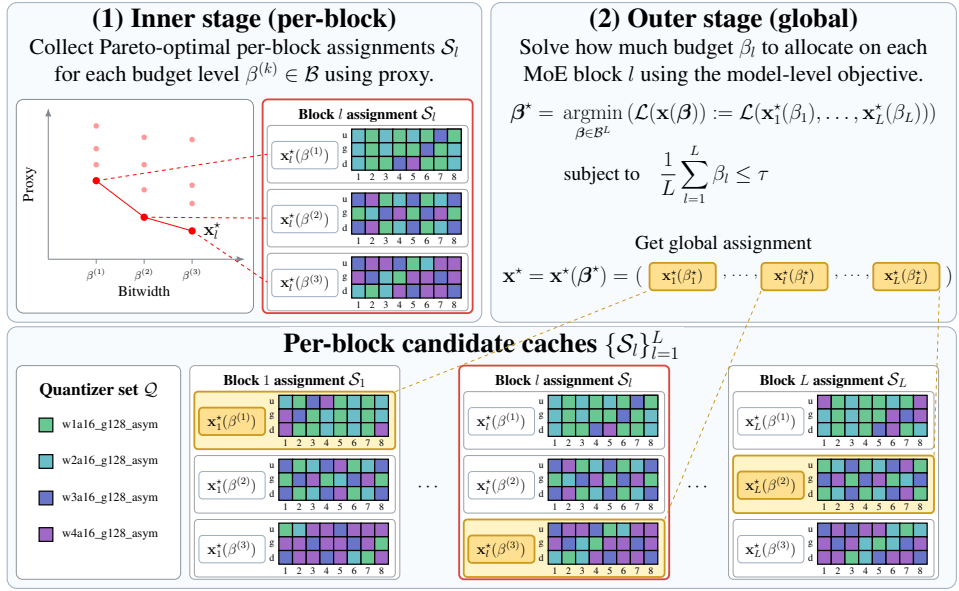
\begin{figure*}[t]
    \centering
    \resizebox{0.8\textwidth}{!}{%
      \begin{tikzpicture}[x=1.3cm,y=1.3cm]
        \LayerSolutionBankFinal
      \end{tikzpicture}%
    }
\caption{Overview of \textsc{Q-Strata}. The inner stage uses the cheap per-block proxy score to cache, for each MoE block, a Pareto frontier of candidate assignments over the budget grid, one assignment per budget level. The outer stage selects one cached candidate per block, that is, one budget $\beta_l$ per block, to minimize the model-level objective $\mathcal{L}$ directly under the average-bitwidth budget $\tau$, then assembles them into the global assignment. This reduces the search from a bitwidth per linear layer to one budget per block.}
\label{fig:overview}
\end{figure*}

Mixture-of-Experts (MoE) models such as Mixtral, DeepSeek-V3, and Qwen3 have become a leading approach for scaling large language models (LLMs) across diverse tasks~\citep{mixtral, deepseekv3, qwen3}.
They route each token to a small subset of experts, which enables them to scale up model capacity without compromising compute efficiency.
The trade-off of this design is a large memory footprint, since all experts must be stored even though only a few are active per token, so reducing this footprint is central to real-world deployment and calls for weight compression that exploits the MoE structure.
In particular, we focus on weight-only post-training quantization (PTQ), which compresses the weights of a trained model with only a small calibration dataset and no retraining, reflecting practical deployment constraints~\citep{gptq, hubara}.

Mixed-precision quantization (MPQ) assigns different bitwidths to the linear layers of a model to minimize the quality lost under a fixed budget, pushing the size-quality frontier of dense models beyond uniform quantization~\citep{qpalette}.
The task is naturally cast as minimizing this quality loss over a high-dimensional discrete space of bitwidth assignments.
Existing methods take one of two routes.
One approximates this loss as a linear sum of layer-wise proxy terms, typically from a Hessian approximation, and solves the allocation with integer linear programming~\citep{hawqv2, chen2021towards}.
The other optimizes a model-level objective, evaluated on the assembled quantized model, in a black-box manner with model-based evolutionary search or reinforcement learning~\citep{amq, apq, haq}.
The proxy route is suboptimal because it optimizes an approximate objective, whereas the black-box route works with the model-level objective directly and reaches higher quality at the cost of repeated model evaluations.

The problem becomes considerably harder for MoE LLMs.
Each of the $L$ MoE blocks holds $E$ experts, and each expert contains three linear layers (gate, up, and down), so the MoE blocks alone hold $3LE$ linear layers, inducing a search space of $|\mathcal{Q}|^{3LE}$ assignments over the $|\mathcal{Q}|$ candidate quantizers, far more than in a dense model.
For instance, within the Qwen3 family, the dense Qwen3-32B has $7 \times 64 = 448$ linear layers, whereas the similarly sized Qwen3-30B-A3B has $3 \times 48 \times 128 = 18{,}432$ linear layers in its MoE blocks alone, more than forty times as many.\footnote{The dense count includes all seven linear layers per block, whereas the MoE count includes only the three projections (up, gate, down) of each of the $128$ experts across the $48$ blocks, excluding attention and routing.}
Under this high dimensionality, black-box search cannot be applied directly.
MxMoE and MC-MoE instead use proxy scores to capture the differing contributions of the experts and linear layers within each MoE block, which improves over uniform quantization, but they keep every block's budget uniform and do not allocate across blocks~\citep{mxmoe, mcmoe}.
GEMQ does allocate different budgets across blocks through a global assignment, but it still relies on an additive proxy for model quality, so it misses the dependencies that quantization creates between blocks~\citep{gemq}.
No existing quantization method for MoE LLMs directly optimizes a model-level objective over the choices that couple the blocks.

Our central observation is that bitwidth allocation comprises two distinct strata.
Within a block, a low-cost proxy ranks candidate assignments reliably, as prior MoE methods demonstrate, whereas across blocks the assignments are coupled, so their joint effect shows up only in a model-level objective evaluated on the assembled quantized model.
To optimize this model-level objective directly without searching the $3LE$-dimensional assignment space, we propose \textsc{Q-Strata}, a bi-level allocator that splits the allocation along these two strata (\Cref{fig:overview}).
Its inner stage caches a Pareto frontier of candidates per block over finely spaced budgets, and its outer stage selects one cached candidate per block to minimize the model-level objective under the global budget, operating over only $L$ per-block budgets rather than $3LE$ individual bitwidths.
We solve this outer selection with a greedy descent that starts from the most expensive budgets and repeatedly lowers the budget of the block whose reduction raises the objective least, made affordable by a lazy evaluation scheme that recomputes only a few marginals per step~\citep{minoux}.
Across Mixtral-8$\times$7B-Instruct~\citep{mixtral}, Qwen1.5-MoE-A2.7B~\citep{qwenmoe}, DeepSeek-V2-Lite~\citep{deepseekv2}, and Qwen3-30B-A3B~\citep{qwen3}, \textsc{Q-Strata} achieves the lowest WikiText2 perplexity among the compared methods in the low-bit regime, and its outer search also serves as a standalone allocator on dense models.

\section{Preliminaries}
\subsection{Problem formulation}
We study weight-only mixed-precision quantization (MPQ) of a Mixture-of-Experts (MoE) model with $L$ MoE blocks indexed by $l\in[L]$, where $[n]=\{1,\dots,n\}$. Each MoE block contains $E$ experts indexed by $e\in[E]$, and each expert holds three linear layers, the up, gate, and down projections of its feed-forward network, indexed by $i\in\mathcal{I}=\{\mathrm{up},\mathrm{gate},\mathrm{down}\}$. A shared expert, when present, is counted as one of the $E$ experts. We write $W_{l,e,i}$ for the full-precision weight of the linear layer $(l,e,i)$, projection $i$ of expert $e$ in MoE block $l$, and $|W_{l,e,i}|$ for its number of parameters. These expert linear layers, $N=3LE$ in total, are the atomic units of bit allocation. The attention projections, the routing gate, and the embeddings hold a small fraction of the parameters and are kept at a fixed precision, so we allocate bits only over the expert linear layers \citep{mxmoe}.

Given a set of candidate quantizers $\mathcal{Q}$, an assignment selects a quantizer $x_{l,e,i}\in\mathcal{Q}$ for each layer, replacing its weight $W_{l,e,i}$ with the quantized weight $x_{l,e,i}(W_{l,e,i})$. We gather these choices into the local assignment $\mathbf{x}_l=(x_{l,e,i})_{e\in[E],\,i\in\mathcal{I}}\in\mathcal{X}_l=\mathcal{Q}^{E\times 3}$ of block $l$ and the global assignment $\mathbf{x}=(\mathbf{x}_1,\dots,\mathbf{x}_L)\in\mathcal{X}=\mathcal{Q}^{L\times E\times 3}$.

The bit cost of block $l$ is $C_l(\mathbf{x}_l)=\sum_{e,i} c_{l,e,i}(x_{l,e,i})$, where $c_{l,e,i}(x)$ is the storage in bits of layer $(l,e,i)$ quantized by $x$, including the scale and zero-point overhead. From this we define the average bitwidth of block $l$, $\bar b_l(\mathbf{x}_l)=C_l(\mathbf{x}_l)\big/\sum_{e,i}|W_{l,e,i}|$, and the average bitwidth over the expert weights, $\bar b(\mathbf{x})=\sum_{l}C_l(\mathbf{x}_l)\big/\sum_{l,e,i}|W_{l,e,i}|$, the total expert-weight bits over the total expert-parameter count. In the MoE models we consider, the blocks are structurally identical and share the same parameter count, so $\bar b(\mathbf{x})$ reduces to the mean of the per-block bitwidths, $\bar b(\mathbf{x})=\tfrac{1}{L}\sum_{l}\bar b_l(\mathbf{x}_l)$.

MPQ minimizes $\mathcal{L}(\mathbf{x})$ at a target average bitwidth $\tau$,
\begin{align}
\label{eq:mpqmain}
& \minimize_{\mathbf{x}\in\mathcal{X}} && \mathcal{L}(\mathbf{x})\\
& ~\mathrm{subject~to} && \bar b(\mathbf{x})\le\tau .\nonumber
\end{align}

$\mathcal{L}(\mathbf{x})$ is a model-level objective that scores the quantized model on the calibration data.
Following the protocol of \citet{amq}, we use the Jensen--Shannon divergence (JSD) between the next-token distributions of the quantized and full-precision models, where a lower value indicates higher quality.
The full-precision distributions are precomputed once, so evaluating $\mathcal{L}$ takes a single end-to-end forward pass of the quantized model on the calibration data.

This optimization is especially hard for MoE models, since the dimension of the assignment space, $N=3LE$, scales with the number of experts rather than with model size or compute.

Rather than optimize over $\mathcal{X}$ monolithically, we exploit the two-level structure of the MoE model.
The allocation inside a block, over its experts and projections, is captured well by a cheap block-local proxy, an estimate that needs no end-to-end forward of the quantized model, whereas the allocation across blocks couples the blocks in a way that only the model-level objective $\mathcal{L}$ reflects.
This contrast lets us split this search into two tractable stages, an inner stage that collects, for each block, a set of bit-proxy trade-off solutions, and an outer stage that optimizes the per-block selection from these sets to minimize   $\mathcal{L}$, the bi-level design we develop in \Cref{sec:method}.

\subsection{Block reconstruction proxy}
For the allocation inside a single MoE block, prior MoE quantization methods rely on the block-output reconstruction error~\citep{mxmoe, mcmoe}.
Let $h_l$ be the output of MoE block $l$ on the calibration inputs and $\hat h_l(\mathbf{x}_l)$ its output after quantizing the block with $\mathbf{x}_l$.
The reconstruction error $\lVert \hat h_l(\mathbf{x}_l)-h_l\rVert^{2}$ depends only on the local assignment $\mathbf{x}_l$, but the gating nonlinearity and routing couple the $3E$ linear layers, so it is not additive across them.
To make it decomposable, these methods approximate it by a proxy that sums the block-output distortion each linear layer induces in isolation,
\[
\mathcal{D}_l(\mathbf{x}_l)=\sum_{e\in[E]}\sum_{i\in\mathcal{I}}\big\lVert \hat h_l^{(e,i)}(x_{l,e,i})-h_l\big\rVert^{2},
\]
where $\hat h_l^{(e,i)}(q)$ is the block output when only linear layer $(l,e,i)$ is quantized with quantizer $q$ and the rest of the block is kept at full precision.
Each distortion is precomputed once per layer and quantizer from a single block forward, and since it is nonzero only on the tokens routed to expert $e$, frequently used experts contribute more, matching their larger effect on the block output.
Two properties of this proxy shape our design.
It is block-separable, since $\mathcal{D}_l$ depends only on $\mathbf{x}_l$ while $\mathcal{L}$ couples all blocks, and it is cheap, since each term needs only a forward of the single block.
We therefore adopt it for the allocation inside each block and reserve $\mathcal{L}$ for the allocation across blocks.

\section{Method}
\label{sec:method}
\textsc{Q-Strata} decomposes \Cref{eq:mpqmain} into two stages that follow the two-level structure of an MoE model (\Cref{fig:overview}).
The inner stage uses the cheap proxy $\mathcal{D}_l$ to precompute, for each MoE block, a small set of Pareto-optimal candidate assignments across budgets, and caches them.
The outer stage optimizes the model-level objective $\mathcal{L}$ directly by choosing one cached candidate per block.
The inner stage decides how to allocate quantizers to linear layers within each block for each budget, and the outer stage decides how to allocate budget across blocks.

\begin{algorithm}[t]
\caption{Inner stage: per-block candidate caching}
\label{alg:inner}
\begin{algorithmic}[1]
% \Require quantizer set $\mathcal{Q}$, budget grid $\mathcal{B}$, full-precision block outputs $\{h_l\}$
\Require quantizer set $\mathcal{Q}$, budget grid $\mathcal{B}$, per-block inputs $\{X_l\}$ and full-precision outputs $\{h_l\}$ from one calibration forward
\Ensure candidate caches $\{\mathcal{S}_l\}_{l=1}^{L}$
\For{$l=1$ to $L$}
  % \For{$(e,i,q)\in[E]\times\mathcal{I}\times \mathcal{Q}$}
  %   \State $d_l(e,i,q)\gets\lVert \hat h_l^{(e,i)}(q)-h_l\rVert^{2}$
  % \EndFor
\For{$(e,i,q)\in[E]\times\mathcal{I}\times \mathcal{Q}$}
    \State $d_l(e,i,q)\gets\lVert \hat h_l^{(e,i)}(q)-h_l\rVert^{2}$
  \EndFor
  \For{$\beta\in\mathcal{B}$}
    \State $\mathbf{x}_l^{\star}(\beta)\gets$ solve \eqref{eq:inner} via ILP
  \EndFor
  \State $\mathcal{S}_l\gets\{\mathbf{x}_l^{\star}(\beta):\beta\in\mathcal{B}\}$
\EndFor
\State \Return $\{\mathcal{S}_l\}_{l=1}^{L}$
\end{algorithmic}
\end{algorithm}

\subsection{Inner stage}
For each MoE block $l$ and budget $\beta$, we select the local assignment that minimizes the proxy under the budget,
\begin{align}
\label{eq:inner}
\mathbf{x}_l^{\star}(\beta)~~&=~~\argmin_{\mathbf{x}_l\in\mathcal{X}_l}~~~~~ \mathcal{D}_l(\mathbf{x}_l)\\
&~~~\mathrm{subject~to~}~~~~~ \bar b_l(\mathbf{x}_l)\le\beta.\nonumber
\end{align}

Because the proxy decomposes over the layers, this is a multiple-choice knapsack problem (MCKP), in which the $3E$ linear layers are the choice groups and the quantizers in $\mathcal{Q}$ are the items, and choosing a quantizer for a layer contributes its bitwidth to the cost and the block-output distortion it induces to the objective.
We solve this MCKP exactly with an integer linear program (ILP) using an off-the-shelf solver~\citep{gurobi}.
The distortion terms $\lVert\hat h_l^{(e,i)}(q)-h_l\rVert^{2}$ summed in $\mathcal{D}_l$ are precomputed once for every layer and quantizer, reusing the block inputs $X_l$ from a single full-precision forward over the calibration set.
We solve over a shared budget grid $\mathcal{B}=\{\beta^{(1)}<\dots<\beta^{(K)}\}$ of $K$ equally spaced levels with step $\Delta=\beta^{(k+1)}-\beta^{(k)}$, and cache the candidates $\mathcal{S}_l=\{\mathbf{x}_l^{\star}(\beta):\beta\in\mathcal{B}\}$. Here, the grid resolution $K$ controls how finely the per-block cost-distortion frontier is sampled.
Caching one candidate per budget level collapses the global search from $\mathcal{Q}^{3LE}$ to a per-block budget choice $\boldsymbol\beta\in\mathcal{B}^{L}$.
Algorithm~\ref{alg:inner} summarizes the inner stage.
\begin{algorithm}[t]
\caption{Outer stage: lazy greedy descent}
\label{alg:outer}
\begin{algorithmic}[1]
\Require caches $\{\mathcal{S}_l\}$, model-level objective $\mathcal{L}$, target average bitwidth $\tau$
\Ensure budgets $\boldsymbol\beta$, assignment $\mathbf{x}(\boldsymbol\beta)$
\State $\beta_l\gets\beta^{(K)}$ for all $l$;\ \ $t\gets 0$;\ \ $\ell\gets\mathcal{L}(\mathbf{x}(\boldsymbol\beta))$
\State $H\gets\emptyset$;\ \ \Call{Refresh}{$l$} for all $l\in[L]$
\While{$\frac{1}{L}\sum_l\beta_l>\tau$ \textbf{and} $H\ne\emptyset$}
  \State pop block $l$ with smallest key $g_l$ from $H$
  \If{$\kappa_l<t$} \Comment{stale}
    \State \Call{Refresh}{$l$}
  \Else
    \State lower $\beta_l$ one level
    \State $\ell\gets\ell+g_l$;\ \ $t\gets t+1$
    \State \textbf{if} $\beta_l>\beta^{(1)}$ \textbf{then} \Call{Refresh}{$l$}
  \EndIf
\EndWhile
\State \Return $\boldsymbol\beta$, $\mathbf{x}(\boldsymbol\beta)$
\Statex
\Function{Refresh}{$l$}
  \State $\boldsymbol\beta'\gets\boldsymbol\beta$ with $\beta_l$ lowered one level
  \State $g_l\gets\mathcal{L}(\mathbf{x}(\boldsymbol\beta'))-\ell$;\ \ $\kappa_l\gets t$
  \State insert $l$ into min-heap $H$ keyed by $g_l$
\EndFunction
\end{algorithmic}
\end{algorithm}

\definecolor{trial01}{HTML}{1F77B4}
\definecolor{trial02}{HTML}{FF7F0E}
\definecolor{trial03}{HTML}{2CA02C}
\definecolor{trial04}{HTML}{D62728}
\definecolor{trial05}{HTML}{9467BD}
\definecolor{trial06}{HTML}{8C564B}
\definecolor{trial07}{HTML}{E377C2}
\definecolor{trial08}{HTML}{7F7F7F}
\definecolor{trial09}{HTML}{BCBD22}
\definecolor{trial10}{HTML}{17BECF}

\subsection{Outer stage}
The outer stage assembles a global assignment from the cached candidates and optimizes the model-level objective $\mathcal{L}$ directly.
Choosing a budget $\beta_l\in\mathcal{B}$ for each block  yields $\mathbf{x}(\boldsymbol\beta)=(\mathbf{x}_1^{\star}(\beta_1),\dots,\mathbf{x}_L^{\star}(\beta_L))$, and we solve
\begin{align}
\label{eq:outer}
\boldsymbol\beta^{\star}~~&=~~\argmin_{\boldsymbol\beta\in\mathcal{B}^{L}}~~~~~ \mathcal{L}\big(\mathbf{x}(\boldsymbol\beta)\big)\\
&~~~\mathrm{subject~to~}~~~~~ \tfrac{1}{L}\textstyle\sum_{l=1}^{L}\beta_l\le\tau,\nonumber
\end{align}
with $\mathbf{x}^{\star}=\mathbf{x}(\boldsymbol\beta^{\star})$.
Unlike the additive proxy of the inner stage, $\mathcal{L}$ does not decompose into a sum over the per-block choices, so this selection cannot be solved as an MCKP.

\subsection{Solving the outer problem}
The outer problem is a black-box combinatorial optimization over the lattice $\mathcal{B}^{L}$ under a budget constraint on the sum of per-block budgets, where each evaluation of $\mathcal{L}$ is a forward pass of the assembled model.

Exhaustive search over $\mathcal{B}^{L}$ is infeasible, and we optimize it with a greedy descent.
Starting from the most expensive corner with $\beta_l=\beta^{(K)}$ for all $l$, we repeatedly lower one block by a single level until the average bitwidth meets $\tau$.
For a block $l$ above the bottom level, let $\boldsymbol\beta^{-l}$ lower $\beta_l$ by one grid level, giving its marginal loss increase $\delta_l(\boldsymbol\beta)=\mathcal{L}(\mathbf{x}(\boldsymbol\beta^{-l}))-\mathcal{L}(\mathbf{x}(\boldsymbol\beta))$.
A budgeted greedy would weigh each move's $\delta_l$ against the budget it saves~\citep{khuller1999}, but here every one-level move saves the same budget, so we lower the block with the smallest $\delta_l$.
Each step evaluates $\mathcal{L}$ on an assembled assignment, so the descent captures the coupling between blocks, but a direct implementation recomputes every $\delta_l$ at each of the $\mathcal{O}(LK)$ steps, $\mathcal{O}(L^{2}K)$ evaluations of $\mathcal{L}$.

% Exhaustive search over $\mathcal{B}^{L}$ is infeasible, and we optimize it with a greedy descent.
% Starting from the most expensive corner with $\beta_l=\beta^{(K)}$ for all $l$, we repeatedly lower one block by a single level until the average bitwidth meets $\tau$.
% A budgeted greedy would weigh each move's increase in $\mathcal{L}$ against the budget it saves~\citep{khuller1999}, but here every one-level move saves the same budget, so we lower the block whose decrease raises $\mathcal{L}$ the least.
% Each step evaluates $\mathcal{L}$ on an assembled assignment, so the descent captures the coupling between blocks, but a direct implementation recomputes all $L$ marginals at each of the $\mathcal{O}(LK)$ steps, $\mathcal{O}(L^{2}K)$ evaluations of $\mathcal{L}$.

\paragraph{Lazy acceleration.}
Our implementation runs this descent with lazy evaluation~\citep{minoux}.
We store each block's last loss increase together with the step at which it was computed, keep them in a min-heap, and at each step pop the smallest, recompute it if stale, and apply the move once the popped value is current.
\Cref{alg:outer} gives the full procedure.

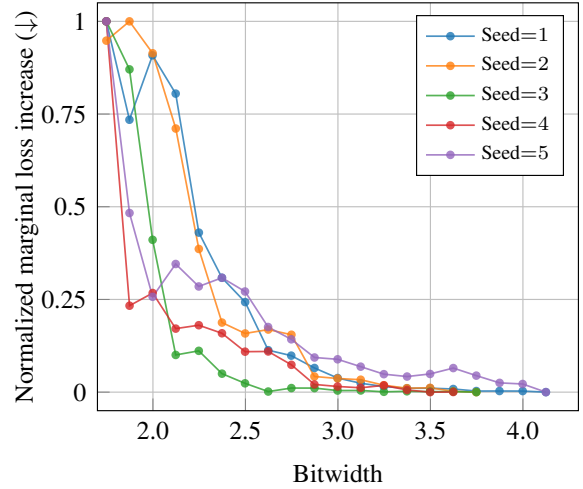
\begin{figure}[t]
    \centering
    \resizebox{\linewidth}{!}{
    \begin{tikzpicture}
    \begin{axis}[
        width=7.5cm,
        height=6.6cm,
        every axis plot/.append style={thick},
        grid=major,
        scaled ticks=false,
        ylabel near ticks,
        tick pos=left,
        tick label style={font=\small},
        xtick={2.0, 2.5, 3.0, 3.5, 4.0},
        xticklabels={ 2.0, 2.5, 3.0, 3.5, 4.0},
        ytick={0, 0.25, 0.5, 0.75, 1},
        yticklabels={0, 0.25, 0.5, 0.75, 1},
        label style={font=\small},
        xlabel={Bitwidth},
        xlabel style={at={(0.5,0)}},
        ylabel={Normalized marginal loss increase ($\downarrow$)},
        ylabel style={align=center, at={(-0.1,0.5)}},
        xmin=1.7, xmax=4.3,
        ymin=-0.05, ymax=1.05,
        legend style={legend columns=1, at={(0.97, 0.97)}, anchor=north east,
                      font=\scriptsize, cells={align=left}},
    ]
    \addplot[trial01, line width=0.6pt, mark=*, mark size=1.2pt, opacity=0.8]
        table[x=bit, y=dr, col sep=comma]{csvs/sanity_dr/ds2_rndm_t1.csv};
    \addlegendentry{Seed${=}1$}
    \addplot[trial02, line width=0.6pt, mark=*, mark size=1.2pt, opacity=0.8]
        table[x=bit, y=dr, col sep=comma]{csvs/sanity_dr/ds2_rndm_t2.csv};
    \addlegendentry{Seed${=}2$}
    \addplot[trial03, line width=0.6pt, mark=*, mark size=1.2pt, opacity=0.8]
        table[x=bit, y=dr, col sep=comma]{csvs/sanity_dr/ds2_rndm_t3.csv};
    \addlegendentry{Seed${=}3$}
    \addplot[trial04, line width=0.6pt, mark=*, mark size=1.2pt, opacity=0.8]
        table[x=bit, y=dr, col sep=comma]{csvs/sanity_dr/ds2_rndm_t4.csv};
    \addlegendentry{Seed${=}4$}
    \addplot[trial05, line width=0.6pt, mark=*, mark size=1.2pt, opacity=0.8]
        table[x=bit, y=dr, col sep=comma]{csvs/sanity_dr/ds2_rndm_t5.csv};
    \addlegendentry{Seed${=}5$}
    \end{axis}
    \end{tikzpicture}}
\caption{Empirical diminishing-returns check on DeepSeek-V2-Lite, the property that lazy descent relies on. Lowering a block by one budget grid level incurs a larger loss increase under stronger compression, increase grows toward the low-bit end. Each curve corresponds to one of five random trials.}
\label{fig:dr}
\end{figure}

This procedure reproduces the eager descent as long as a stale loss increase only underestimates its current value, so that the min-heap never hides the least harmful move. Our $\mathcal{L}$ carries no guarantee of this condition, so we use lazy evaluation as a heuristic, justified empirically by the near-monotone growth of the marginals along the descent (\Cref{fig:dr}). \Cref{def:dr} states the diminishing-returns property under which the condition holds exactly.

\begin{definition}[DR-submodularity on the budget grid]
\label{def:dr}
For $\boldsymbol\beta\in\mathcal{B}^{L}$ and a block $l$ below the top level, let $\boldsymbol\beta^{+l}$ raise $\beta_l$ to the next grid level. A function $G:\mathcal{B}^{L}\to\mathbb{R}$ is DR-submodular if, for all $\boldsymbol\beta\le\boldsymbol\beta'$ taken coordinatewise and every such block $l$,
\[
G(\boldsymbol\beta^{+l})-G(\boldsymbol\beta)\ \ge\ G\big((\boldsymbol\beta')^{+l}\big)-G(\boldsymbol\beta'),
\]
so the gain from raising a block by one level does not increase as the other budgets grow~\citep{somayoshida}.
\end{definition}
When $G=-\mathcal{L}$ satisfies \Cref{def:dr}, the descent only lowers budgets, so a $\delta_l$ cached at a higher budget can only be exceeded later, exactly the underestimate condition above.
\begin{table*}[t]
\centering
\resizebox{\textwidth}{!}{%
\begin{tabular}{l ccc ccc ccc}
\toprule
& \multicolumn{3}{c}{\makecell{Mixtral-8$\times$7B \\ (46.7B - A12.9B)}} 
& \multicolumn{3}{c}{\makecell{Qwen1.5-MoE \\ (14.3B - A2.7B)}} 
& \multicolumn{3}{c}{\makecell{DeepSeek-V2-Lite \\ (15.7B - A2.4B)}} \\
\cmidrule(lr){2-4} \cmidrule(lr){5-7} \cmidrule(lr){8-10}
Method
& Bits ($\downarrow$) & Wiki2 ($\downarrow$) & Acc ($\uparrow$)
& Bits ($\downarrow$) & Wiki2 ($\downarrow$) & Acc ($\uparrow$)
& Bits ($\downarrow$) & Wiki2 ($\downarrow$) & Acc ($\uparrow$) \\
\midrule
BF16       & 16.00 & 3.88 & 81.25 & 16.00 & 6.79 & 69.98       & 16.00 & 5.92 & 72.29 \\
\midrule
% 2.25
GPTQ   & 2.25 & 5.73 & 70.89 & 2.25 & 10.41 & 55.73   & 2.25 & 8.59 & 60.81\\
MxMoE & 2.25 & 5.69 & 71.70  & 2.25 & 8.07 & 61.93  & 2.25 & 6.78 & \textbf{64.04} \\
GEMQ (shared)   & 2.25 &   9.39   &  63.63     & 2.25 &   10.76      &  58.67           & 2.25 &  7.61     &  60.21  \\
\textsc{Q-Strata}   & 2.25 & \textbf{5.62} & \textbf{71.78} & 2.25 & \textbf{7.98} & \textbf{62.93}  & 2.25 & \textbf{6.74} & 63.75 \\
\midrule
% 2
GPTQ$^{\dagger}$    & 2.01 & 11.37 & 44.44  & 2.02 & 18.64 & 43.24 & 2.02 & 13.79 & 47.98\\
MxMoE   & 2.00 & 8.03 & 61.42& 2.00 & 9.06 & 57.16   & 2.00 & 7.46 & 59.73\\
GEMQ (shared)    & 2.00 &  19.21  &  53.30  & 2.00 & 15.14    & 53.15      & 2.00 &  10.28   &  53.36   \\
\textsc{Q-Strata}    & 2.00 & \textbf{6.90} & \textbf{63.79}& 2.00 & \textbf{8.97} & \textbf{58.41}   & 2.00 & \textbf{7.41} & \textbf{59.88}\\
\midrule
% 1.75
MxMoE   & 1.75 & 25.20 & 44.58 & 1.75 & 12.49 & 52.38  & 1.75 & 9.90 & 50.95\\
GEMQ (shared)     & 1.75 & 21.25 & 49.40   & 1.75 &  23.89    & 48.93      & 1.75 & 19.34     &   44.85   \\
\textsc{Q-Strata}     & 1.75 & \textbf{12.14} & \textbf{52.51}& 1.75 & \textbf{11.84} & \textbf{53.74} & 1.75 & \textbf{9.35} & \textbf{52.46} \\
\bottomrule
\end{tabular}
}
\caption{Mixed-precision quantization results on MoE LLMs across target bitwidths. All methods share the same weight-only GPTQ quantizer set (1,2,3,4\,bit, group size 128, asymmetric). GPTQ, MxMoE, and \textsc{Q-Strata} use the rotated setting of \Cref{sec:setup}, while GEMQ (shared) deploys GEMQ's allocation in its original rotation-free setting without router fine-tuning. \Cref{sec:gemq} compares against GEMQ with the router fine-tuning of its official recipe restored, with and without rotation. $^{\dagger}$The $2.0$-bit uniform GPTQ rows use the per-channel 2-bit format with $2.0$ effective bits, since no uniform group-128 configuration meets that budget.}
\label{tab:quant}
\end{table*}

\begin{remark}
\label{prop:lazy}
If $G=-\mathcal{L}$ is DR-submodular (\Cref{def:dr}), this lazy descent commits the same moves as the eager descent that recomputes all $L$ marginals each step, while evaluating $\mathcal{L}$ fewer times \citep{minoux,somayoshida}.
\end{remark}
Empirically, $\mathcal{L}$ is close to this property. A more compressed model is more fragile, so lowering one more bitwidth level hurts quality more once few bits remain. We check this on DeepSeek-V2-Lite by following random descent paths and recording, one block at a time, the resulting loss increase $\delta_l$ from a single bitwidth level (\Cref{fig:dr}). This $\delta_l$, which the heap caches, grows near-monotonically as the model is compressed, so a value cached at a higher budget typically underestimates its current one, the condition the lazy descent relies on. \Cref{app:dr} gives the protocol.

\paragraph{Cost.}
The outer search's cost is its evaluations of $\mathcal{L}$, each a forward pass of the assembled model.
The descent lowers budgets monotonically and uses $\tau$ only as a stopping point, so a single top-to-bottom sweep yields the assignment for every target budget along the way.
Writing $R$ for the average number of evaluations per committed move, the lazy descent issues $\mathcal{O}(RLK)$, smaller than the eager $\mathcal{O}(L^{2}K)$ by a factor $L/R$.
In practice, $R$ stays near $2.5$, far below the $24$ to $32$ blocks (\Cref{tab:efficiency} in \Cref{app:dr}).

\section{Experiments}
\label{sec:exp}

\subsection{Setup}
\label{sec:setup}
We mainly evaluate three MoE LLMs of different scales, Mixtral-8$\times$7B-Instruct (46.7B - A12.9B), Qwen1.5-MoE (14.3B - A2.7B), and DeepSeek-V2-Lite (15.7B - A2.4B) \citep{mixtral, qwenmoe,deepseekv2}, and scale the comparison up to Qwen3-30B-A3B~\citep{qwen3} in \Cref{app:scale}.
All compared methods quantize weights with GPTQ~\citep{gptq} and share one quantizer set, the group-128 asymmetric formats at $1$, $2$, $3$, and $4$ bits.
The budget grid spans $1.25$ to $4.25$ bits in steps of $0.125$, giving $K=25$ levels.
We report WikiText2 perplexity and the average accuracy over six zero-shot tasks, PIQA, BoolQ, WinoGrande, ARC-easy, ARC-challenge, and HellaSwag \citep{piqa,boolq,winogrande,arc,hellaswag}.

The baselines are uniform GPTQ, MxMoE, and GEMQ.
Uniform GPTQ gives every linear layer the same bitwidth \citep{gptq}.
MxMoE shares our inner stage but assigns every MoE block the same budget $\tau$ rather than optimizing per-block budgets, making it the counterpart that omits our outer stage \citep{mxmoe}.
GEMQ allocates bitwidth globally at expert granularity, solving an ILP that minimizes a gradient-based additive proxy for model quality \citep{gemq}.
Uniform GPTQ, MxMoE, and \textsc{Q-Strata} apply the quantizers after a random Hadamard rotation without online rotation, following the protocol of \citet{mxmoe} \citep{quarot}, so differences among these three reflect bit allocation alone.
GEMQ is reproduced under a shared protocol, labeled GEMQ (shared) in \Cref{tab:quant}, by taking its allocation from its own gradient-proxy ILP and deploying it with the shared quantizer set, keeping its original rotation-free setting and omitting its router fine-tuning (RFT) step, an additional training stage.\footnote{GEMQ's original pipeline further differs from the shared protocol in its GPTQ variant (MSE range search vs.\ min-max), its 1-bit format (symmetric, 1.125 effective bits, vs.\ asymmetric, 1.25), and its evaluation context length (2048 vs.\ our 4096).}
\Cref{sec:gemq} additionally compares \textsc{Q-Strata} against GEMQ with the RFT of its official recipe restored, with and without rotation.
Full details, including the per-stage calibration sizes and the quantization scope shared by all mixed-precision methods, are in \Cref{app:setup}.

\subsection{Main results}
\Cref{tab:quant} reports results at target bitwidths $2.25$, $2.00$, and $1.75$.
\textbf{\textsc{Q-Strata} attains the lowest WikiText2 perplexity for every model at every bitwidth, and the best zero-shot accuracy in all cases but one}, DeepSeek-V2-Lite at $2.25$ bits, where MxMoE leads by $0.3$ points.
At $1.75$ bits on Mixtral, \textsc{Q-Strata} more than halves the perplexity of MxMoE, $12.14$ against $25.20$, and improves average accuracy by almost eight points.
Because MxMoE is exactly \textsc{Q-Strata} without the outer stage (\Cref{sec:setup}), \Cref{tab:quant} decomposes the two stages.
The GPTQ-to-MxMoE gap isolates the inner stage's gain, and the MxMoE-to-\textsc{Q-Strata} gap isolates the outer stage's gain, which widens as the budget tightens.
The gains also persist on the larger Qwen3-30B-A3B with $18{,}432$ expert linear layers, at a one-time search cost of 23 to 89 GPU-hours per model (\Cref{app:scale}).
Further analyses of the fidelity of the calibration objective, calibration-subset stability, downstream MMLU and GSM8K accuracy, and the representativeness of the calibration routing are in \Cref{app:corr,app:seeds,app:mmlu,app:routing}.

\begin{table}[t]
\centering
\resizebox{\columnwidth}{!}{
\begin{tabular}{ll rrr}
\toprule
Bits & Outer & JSD\,($\downarrow$) & Wiki2\,($\downarrow$)\\
\midrule
\multirow{3}{*}{2.75}
  & Uniform        & 121.2 & 4.72 \\
  & One-shot ILP  & 108.2 & 4.64 \\
  & Ascending lazy greedy  & 243.9 &  5.62 \\
  & Lazy greedy (\textsc{Q-Strata})  & \textbf{107.2} & \textbf{4.61} \\

\midrule
\multirow{3}{*}{2.25}
  & Uniform        & 234.3 & 5.69 \\
  & One-shot ILP  & 231.6 & 5.64 \\
  & Ascending lazy greedy  &733.6 & 13.22\\
  & Lazy greedy (\textsc{Q-Strata})   & \textbf{229.6} & \textbf{5.62}\\
\midrule
\multirow{3}{*}{1.75}
  & Uniform        & 1036.5 & 25.20\\
  & One-shot ILP  & 823.4 & 13.72 \\
  & Ascending lazy greedy  &1195.9  & 32.76\\
  & Lazy greedy (\textsc{Q-Strata})   & \textbf{749.4} & \textbf{12.14} \\
\bottomrule
\end{tabular}
}
\caption{Outer-stage ablation on Mixtral-8$\times$7B-Instruct with the inner ILP fixed. JSD is the model-level objective $\mathcal{L}$ on the calibration set. }
\label{tab:ablation}
\end{table}

\subsection{Outer-stage ablation}

\Cref{tab:ablation} isolates the outer stage on Mixtral-8$\times$7B-Instruct with the inner ILP fixed, comparing four ways to assign the per-block budgets.
Uniform gives every block the same budget and coincides with the MxMoE baseline.
One-shot ILP scores each block in isolation, recording how much lowering one block to a given level raises $\mathcal{L}$ with the others kept at the highest budget level. It sums these per-block costs into a separable proxy for $\mathcal{L}$ and optimizes it exactly with an ILP in $(K-1)L+1$ evaluations, the search of \citet{qpalette}.
Because this proxy is separable, it ignores the coupling that arises when several blocks are compressed together.
Ascending greedy runs our descent in reverse, raising budgets from the cheapest corner, and trails even the uniform budget.
\textbf{Our lazy greedy outer method attains the lowest JSD, the objective that the outer stage optimizes, and the lowest perplexity at every bitwidth.}
The gap tracks the budget, it becomes large at $1.75$ bits, where One-shot ILP trails our descent by $74$ JSD. 
This shortfall reflects that coupling, which our descent captures by evaluating $\mathcal{L}$ directly at every step.
We define each baseline in more details in \Cref{app:baselines}.
\suppressfloats[t]%
\begin{table}[t]
\centering
\resizebox{\columnwidth}{!}{
\begin{tabular}{ll rr}
\toprule
Bits & Method & JSD\,($\downarrow$) & Wiki2\,($\downarrow$)\\
\midrule
\multirow{3}{*}{2.25}
  & Direct cell-ILP (pruning-free) & \textbf{109.6} & \textbf{6.74} \\
  & One-shot ILP & 112.5 & 6.75 \\
  & Lazy greedy (\textsc{Q-Strata}) & 111.9 & \textbf{6.74} \\
\midrule
\multirow{3}{*}{2.00}
  & Direct cell-ILP (pruning-free) & \textbf{184.4} & \textbf{7.41} \\
  & One-shot ILP & 190.9 & 7.44 \\
  & Lazy greedy (\textsc{Q-Strata}) & 190.0 & \textbf{7.41} \\
\midrule
\multirow{3}{*}{1.75}
  & Direct cell-ILP (pruning-free) & 410.6 & 10.03 \\
  & One-shot ILP & 419.2 & 9.92 \\
  & Lazy greedy (\textsc{Q-Strata}) & \textbf{364.0} & \textbf{9.35} \\
\bottomrule
\end{tabular}}
\caption{Pruning-free direct cell-level ILP, which needs no inner stage but more than $8\times$ the model-level evaluations of our entire search, against two \textsc{Q-Strata} variants that share the inner-stage pruning, on DeepSeek-V2-Lite.}
\label{tab:cellilp}
\end{table}

\begin{table*}[t]
\centering
\resizebox{\textwidth}{!}{%
\begin{tabular}{ll cccc cccc}
\toprule
& & \multicolumn{4}{c}{w/o rotation} & \multicolumn{4}{c}{w/ rotation} \\
\cmidrule(lr){3-6} \cmidrule(lr){7-10}
Model & Bits
& \makecell{\textsc{Q-Strata}\\w/o RFT} & \makecell{\textsc{Q-Strata}\\w/ RFT} & \makecell{GEMQ\\w/o RFT} & \makecell{GEMQ\\w/ RFT}
& \makecell{\textsc{Q-Strata}\\w/o RFT} & \makecell{\textsc{Q-Strata}\\w/ RFT} & \makecell{GEMQ\\w/o RFT} & \makecell{GEMQ\\w/ RFT} \\
\midrule
\multirow{3}{*}{Mixtral-8$\times$7B}
 & 2.25 & 5.61 & \textbf{5.56} & 9.39 & 6.88 & \textbf{5.62} & 5.78 & 20.26 & 7.14 \\
 & 2.00 & 6.78 & \textbf{6.59} & 19.21 & 8.65 & \textbf{6.90} & 7.17 & 46.11 & 8.70 \\
 & 1.75 & 10.22 & \textbf{8.92} & 21.25 & 12.40 & 12.14 & \textbf{9.88} & 99.17 & 15.47 \\
\midrule
\multirow{3}{*}{Qwen1.5-MoE}
 & 2.25 & 8.03 & \textbf{7.75} & 10.76 & 8.91 & 7.98 & \textbf{7.74} & 11.23 & 9.00 \\
 & 2.00 & 9.10 & \textbf{8.66} & 15.14 & 11.05 & 8.97 & \textbf{8.61} & 15.70 & 10.89 \\
 & 1.75 & 12.38 & \textbf{11.47} & 23.89 & 15.14 & 11.84 & \textbf{10.97} & 23.38 & 14.43 \\
\midrule
\multirow{3}{*}{DeepSeek-V2-Lite}
 & 2.25 & 6.80 & \textbf{6.66} & 7.61 & 6.99 & 6.74 & \textbf{6.67} & 8.13 & 7.14 \\
 & 2.00 & 7.42 & \textbf{7.20} & 10.28 & 8.30 & 7.41 & \textbf{7.25} & 12.06 & 8.72 \\
 & 1.75 & 9.36 & \textbf{8.80} & 19.34 & 12.19 & 9.35 & \textbf{8.79} & 23.57 & 13.01 \\
\bottomrule
\end{tabular}}
\caption{WikiText2 perplexity of \textsc{Q-Strata} and GEMQ with and without rotation and with and without GEMQ's router fine-tuning (RFT), applied equally to both methods' allocations under the quantizer set and GPTQ options of \Cref{tab:quant}, at context length 4096. Note that GEMQ's rotated allocation is not a post hoc transfer, as its statistics are recomputed on the rotated model and its ILP re-solved.}
\label{tab:gemq}
\end{table*}
\paragraph{Does the inner-stage pruning lose candidates?}
The outer stage can only pick from the proxy-pruned frontiers, so we compare against a baseline that prunes nothing.
For every (expert, linear layer, bitwidth) cell, this baseline directly measures the model-level JSD change of assigning that cell each candidate bitwidth, then solves the global assignment over raw cells as a single ILP, with no inner stage and no frontier pruning, at about $15.4$k end-to-end evaluations on DeepSeek-V2-Lite, more than $8\times$ the model-level evaluations of \textsc{Q-Strata}'s entire search (\Cref{tab:cellilp}, details in \Cref{app:baselines}).
Comparing this baseline against our one-shot ILP outer stage isolates the effect of the pruning and the block decomposition, and the gap stays within about 4\% in JSD with comparable perplexity at every budget, so the inner stage prunes little that the pruning-free search could exploit.
The remaining $1.75$-bit gap is instead attributable to the search strategy, as the descent reaches JSD $364$ where both one-shot solvers, including the pruning-free one, stall at $410$ to $419$, again reflecting the inter-block coupling captured during the descent.

\subsection{Router fine-tuning and rotation}
\label{sec:gemq}
The shared protocol of \Cref{tab:quant} removes GEMQ's router fine-tuning, so we restore it.
RFT following GEMQ's official recipe (\Cref{app:setup}) is applied equally to the allocations of GEMQ and \textsc{Q-Strata} at all 9 (model, budget) cells, and both methods are evaluated with and without the Hadamard rotation (\Cref{tab:gemq}).
Notably, \textsc{Q-Strata} attains lower perplexity than GEMQ in all 18 pairs.

\Cref{tab:gemq} also shows that the rotated setting, inherited from MxMoE's pipeline, is not always the better one, since the low budgets force many expert linears to 1 bit, where the effect of rotation becomes model-dependent.
In particular, the strongest Mixtral results at the lowest budgets come from the rotation-free setting, and \Cref{app:rotation} gives the full analysis of how the effect of rotation varies across models.

\subsection{The outer search on dense models}
\label{sec:dense}
The outer search does not rely on the MoE structure and works as a standalone allocator.
On the dense Llama-2-7B we apply it per linear layer rather than per block.
Because the layers differ in size, a one-level move no longer saves the same budget, so the descent instead weighs each move's loss increase against the bits it saves.
This is the budgeted greedy of \citet{khuller1999}, which we detail in \Cref{app:dense}.

We use the HQQ quantizer set (2, 3, and 4 bits, group size 128, asymmetric) and, as in the MoE experiments, take the search objective to be the JSD between the quantized and full-precision models.
We compare our lazy greedy outer allocator against the state-of-the-art black-box optimizer AMQ~\citep{amq}, One-shot ILP, and, for the small calibration setting with 2K tokens, the eager greedy that recomputes every marginal each step.

\begin{table}[t]
\centering
\resizebox{0.94\columnwidth}{!}{
\begin{tabular}{l c c c}
\toprule
Method & Bits & JSD ($\downarrow$) & Queries ($\downarrow$)\\
\midrule
\multirow{3}{*}{One-shot ILP}       & 3.5 & 0.0233 & \multirow{3}{*}{449}\\
                                    & 3.0 & 0.0587 & \\
                                    & 2.5 & 0.1984 & \\
\midrule
\multirow{3}{*}{AMQ}                & 3.5 & 0.0229 & \multirow{3}{*}{10{,}474}\\
                                    & 3.0 & 0.0550 & \\
                                    & 2.5 & 0.1572 & \\
\midrule
\multirow{3}{*}{Lazy greedy}        & 3.5 & 0.0228 & \multirow{3}{*}{1{,}378}\\
                                    & 3.0 & 0.0549 & \\
                                    & 2.5 & 0.1569 & \\
\midrule
\multirow{3}{*}{Eager greedy}       & 3.5 & 0.0224 & \multirow{3}{*}{70{,}345}\\
                                    & 3.0 & 0.0539 & \\
                                    & 2.5 & 0.1568 & \\
\bottomrule
\end{tabular}}
\caption{Search-method comparison on the dense Llama-2-7B with the HQQ quantizer set (2,3,4 bit, group size 128, asymmetric) and $2$K calibration tokens, across average bitwidths $2.5$ to $3.5$. Lazy greedy, the search used in our outer stage, nearly matches the eager greedy at far fewer queries.}
\label{tab:dense-search}
\end{table}

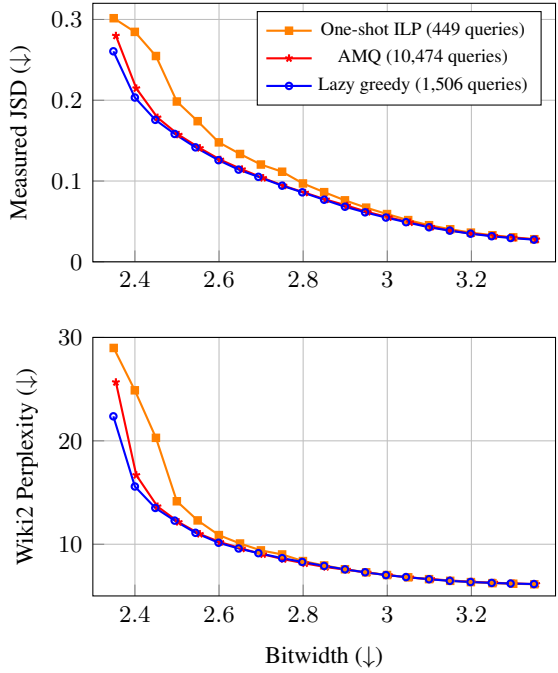
\begin{figure}[t]
\centering
\begin{tikzpicture}
\begin{groupplot}[
    group style={group size=1 by 2, vertical sep=1.0cm},
    width=\columnwidth, height=5cm,
    every axis plot/.append style={thick},
    grid=major, scaled ticks=false,
    ylabel near ticks, tick pos=left,
    tick label style={font=\small}, label style={font=\small},
    xtick={2.4,2.6,2.8,3.0,3.2}, xmin=2.3, xmax=3.4,
]
% --- Top: measured JSD ---
\nextgroupplot[
    ylabel={Measured JSD ($\downarrow$)}, ymin=0, ymax=0.32,
    legend style={at={(0.97,0.97)}, anchor=north east, font=\scriptsize, cells={align=right}},
]
    \addplot[orange, mark size=1.2pt, mark=square*] table [x=bit, y=jsd, col sep=comma]{csvs/hqq-hqq/oneshot_lp.csv};
    \addlegendentry{One-shot ILP (449 queries)}
    \addplot[red, mark size=1.5pt, mark=star] table [x=bit, y=jsd, col sep=comma]{csvs/hqq-hqq/amq.csv};
    \addlegendentry{AMQ (10,474 queries)}
    \addplot[blue, mark size=1.2pt, mark=o] table [x=bit, y=jsd, col sep=comma]{csvs/hqq-hqq/lazy_greedy.csv};
    \addlegendentry{Lazy greedy (1,506 queries)}
% --- Bottom: Wiki2 PPL ---
\nextgroupplot[
    xlabel={Bitwidth ($\downarrow$)}, ylabel={Wiki2 Perplexity ($\downarrow$)},
    ymin=5, ymax=30,
]
    \addplot[red, mark size=1.5pt, mark=star] table [x=bit, y=ppl, col sep=comma]{csvs/hqq-hqq/amq.csv};
    \addplot[orange, mark size=1.2pt, mark=square*] table [x=bit, y=ppl, col sep=comma]{csvs/hqq-hqq/oneshot_lp.csv};
    \addplot[blue, mark size=1.2pt, mark=o] table [x=bit, y=ppl, col sep=comma]{csvs/hqq-hqq/lazy_greedy.csv};
\end{groupplot}
\end{tikzpicture}
\caption{Search-method comparison on Llama 2-7B with the HQQ quantizer set (2,3,4 bit, group size 128, asymmetric) and $8$K calibration tokens. The top panel reports measured JSD on the calibration set and the bottom panel WikiText2 test perplexity. Lazy greedy matches or beats AMQ on both at a fraction of the objective queries listed in the legend with 6.9$\times$ fewer queries.}
\label{fig:hqq_l27b_search_cmpr}
\end{figure}

% We use the HQQ quantizer set (2,3,4 bit, group size 128, asymmetric) and utilize JSD between quantized model and the full precision model as the quality score as in MoE experiments during MPQ search.
% We compare our lazy greedy outer allocator against the state-of-the-art black-box optimizer AMQ~\citep{amq}, One-shot ILP, and, for the small calibration data setting with 2$K$ tokens, the full greedy that recomputes every marginal each step.
\Cref{tab:dense-search} reports the comparison at $2$K calibration tokens.
Lazy greedy matches or beats AMQ at every budget while issuing far fewer queries, $1{,}378$ against $10{,}474$.
Eager greedy is best, and lazy greedy stays within $0.001$ JSD of it while using $51\times$ fewer queries, $1{,}378$ against $70{,}345$, so it closely follows the eager greedy solution.
\Cref{fig:hqq_l27b_search_cmpr} extends the comparison across budgets at $8$K calibration tokens.
\textbf{Surprisingly, our lazy greedy outer method matches or beats AMQ on both while issuing $6.9\times$ fewer queries. }

\section{Related Work}
\label{sec:related}
MPQ is often cast as a multiple-choice knapsack problem~\citep{chen2021towards,mckp,higgs,qpalette}. A line of work approximates the loss with an additive proxy, a sum of per-layer importance scores estimated either from the backward signal~\citep{hawq,hawqv2,hawqv3,lampq,gemq} or from the loss change under full forward passes~\citep{higgs,qpalette}, and solves the allocation with dynamic programming or integer programming~\citep{mckpdp,gurobi}. For dense models, AMQ runs an AutoML-style search with search-space pruning and a model-based evolutionary method, scoring candidates with full forward passes~\citep{amq}.
On the activation side, TWLA's inter-layer-aware ILA-AMP allocates activation bitwidths across layers~\citep{twla}, a complementary axis to the weight-only allocation we study.

For MoE models, prior methods score within-block assignments with block-local proxies. MC-MoE assigns a bitwidth to every expert with a linear program over a hand-designed score combining reconstruction error and routing statistics~\citep{mcmoe}, and MxMoE works at the finer linear-layer granularity, minimizing a sum of block-wise reconstruction-error proxies~\citep{mxmoe}, but both keep every block's budget uniform. GEMQ allocates across blocks through an additive proxy resting on a first-order approximation of the loss~\citep{gemq}. Our inner stage keeps MxMoE's linear-layer granularity, while our outer stage optimizes the model-level objective directly across blocks.

In concurrent work, ScaleBits~\citep{scalebits} quantizes dense models at the tile level. It overlaps with us in a diminishing-returns property over its search space and a greedy allocator, but its space is far larger and finer than our per-block budgets, so it ranks moves by sensitivity-based estimates rather than by measured marginals of its objective. Our bi-level hierarchy shrinks the outer space enough to run the greedy on the objective's marginals directly, and we use lazy evaluation only to cut redundant queries.

\section{Conclusion}
We introduced \textsc{Q-Strata}, a bi-level allocator for mixed-precision quantization of MoE LLMs.
Its inner stage ranks within-block assignments with a cheap proxy and caches one candidate per budget level, so the outer stage only needs to choose one budget per block.
This reduction lets the outer stage optimize the model-level objective directly with a lazy greedy descent, capturing the coupling between blocks that additive proxies miss.
On three MoE LLMs, \textsc{Q-Strata} achieves lower WikiText2 perplexity than uniform GPTQ, MxMoE, and GEMQ in the low-bit regime, and the gains over MxMoE persist on the larger Qwen3-30B-A3B.
The outer search also works as a standalone allocator on dense models, matching the black-box search of AMQ with far fewer objective queries.

\section*{Limitations}
The outer stage evaluates the model-level objective directly, so each move is a forward pass of the assembled model.
Even with lazy evaluation, this issues more such evaluations than allocators that fit an additive proxy once and then solve it in closed form.
The outer search's evaluation count grows with the block count $L$ rather than with the total number of linear layers $3LE$, since the evaluations per move stay near 2.5 and far below $L$ (\Cref{tab:efficiency}).
Because it runs once before deployment and adds nothing to inference latency, its cost is amortized over all subsequent inferences.
It nonetheless remains heavier than proxy-only allocation at search time, and lowering it further is left to future work.

\section*{Acknowledgments}
This work was supported by Samsung Electronics Co., Ltd. (IO250418-12669-01), % NPRC
Mobile eXperience (MX) Business, Samsung Electronics Co., Ltd., % MX
Institute of Information \& Communications Technology Planning \& Evaluation (IITP) grant funded by the Korea government (MSIT)
[No. RS-2026-25524173, Ultra-Long-Term Hierarchical Memory and Reasoning Architecture for Next-Generation Omnimodal Agents, 40\%;
No. RS-2020-II200882, (SW STAR LAB) Development of deployable learning intelligence via self-sustainable and trustworthy machine learning, 10\%;
No. RS-2026-25522672, Development of Unified Reasoning Technology Mimicking Human Cognition for Hierarchical Understanding and Unbounded Problem Solving, 10\%;
and No. RS-2021-II211343, Artificial Intelligence Graduate School Program (Seoul National University), 10\%], % Omnimodal, STAR LAB, human-mimick, AI graduate
and  National Research Foundation of Korea (NRF) grant funded by the Korea government (MSIT) (No. RS-2024-00354036, 30\%). % NRF-Testing
Hyun Oh Song is the corresponding author.

\bibliography{custom}

\appendix
\section{Empirical diminishing returns}
\label{app:dr}
The lazy descent (\Cref{prop:lazy}) is exact when the per-block marginal loss increase $\delta_l(\boldsymbol\beta)=\mathcal{L}(\mathbf{x}(\boldsymbol\beta^{-l}))-\mathcal{L}(\mathbf{x}(\boldsymbol\beta))$ does not shrink as the model is compressed, equivalently when the marginal gain of $G=-\mathcal{L}$ from restoring one level diminishes as the model grows more precise, which is the condition of \Cref{def:dr}.
We probe this on DeepSeek-V2-Lite, following ScaleBits~\citep{scalebits}, which verifies the same property for MPQ at a different granularity.

For each of five trials we sample one descent chain and one block $l$.
A chain starts at the top corner $\boldsymbol\beta=(\beta^{(K)},\dots,\beta^{(K)})$ and lowers a uniformly random block by one level at a time down to the bottom corner, a monotone path along which the effective bitwidth $\bar b(\boldsymbol\beta)$ decreases.
At each configuration along the chain where block $l$ is below the top level, we raise $\beta_l$ by one level and record the marginal loss increase $\mathcal{L}(\mathbf{x}(\boldsymbol\beta))-\mathcal{L}(\mathbf{x}(\boldsymbol\beta^{+l}))$ that its current level carries relative to one level up, on the calibration set used by the outer stage.
We min-max normalize this quantity within each chain and plot it against $\bar b(\boldsymbol\beta)$ in \Cref{fig:dr}.

All five curves rise near-monotonically as the bitwidth falls, so this marginal grows as the model is compressed.
The loss increase $\delta_l$ the lazy heap caches therefore grows along the descent, so a value cached earlier at a higher budget underestimates its value later, exactly what the heap needs.
$\mathcal{L}$ is not strictly DR-submodular and a few small reversals appear, but the trend is monotone enough for the cached values to serve as the underestimates the lazy heap relies on, at the per-move cost reported in \Cref{tab:efficiency}.

\begin{table}[t]
\centering
\resizebox{\columnwidth}{!}{
\begin{tabular}{l ccc}
\toprule
& \makecell{Mixtral\\8$\times$7B} & \makecell{Qwen1.5\\MoE} & \makecell{DeepSeek\\V2-Lite}\\
\midrule
Blocks $L$    & 32 & 24 & 26\\
Levels $K$    & 25 & 25 & 25\\
Commits       & 768 & 576 & 624\\
Lazy evals    & 1{,}909 & 1{,}552 & 1{,}667\\
Eager evals   & 22{,}327 & 13{,}537 & 15{,}802\\
$R$ per move  & 2.49 & 2.69 & 2.67\\
Speedup       & 11.7$\times$ & 8.7$\times$ & 9.5$\times$\\
\bottomrule
\end{tabular}}
\caption{Cost of the outer lazy greedy descent over a full top-to-bottom sweep ($L(K-1)$ commits). $R$ is the average number of objective evaluations per committed move, and Speedup is the ratio of eager to lazy evaluations. Eager evals is the count an eager descent would perform along the same trace, computed from the lazy run rather than from a separate eager execution. }
\label{tab:efficiency}
\end{table}

\section{Experimental details}
\label{app:setup}
\paragraph{Models and quantization.}
We use Mixtral-8$\times$7B-Instruct, Qwen1.5-MoE-A2.7B, and DeepSeek-V2-Lite, and, for \Cref{tab:qwen3}, Qwen3-30B-A3B under the same protocol.
Weights are quantized with GPTQ~\citep{gptq} from the quantizer set of group-128 asymmetric weight-only formats at $1$, $2$, $3$, and $4$ bits.
In the rotated setting, a random Hadamard rotation is applied offline with no online rotation, following \citet{mxmoe}.
The budget grid runs from $1.25$ to $4.25$ bits in steps of $0.125$, so $K=25$.
For all MoE models, we do not quantize linear layers outside of MoE blocks.

\begin{table}[t]
\centering
\resizebox{\columnwidth}{!}{
\begin{tabular}{l ccc}
\toprule
Component & Mixtral & Qwen1.5-MoE & DSv2-Lite\\
\midrule
Expert FFN linears & mixed & mixed & mixed \\
Shared expert & --- & mixed & mixed \\
Routing gates & 16-bit & 16-bit & 16-bit \\
Attention linears & 16-bit & 16-bit & 16-bit \\
Dense MLP (layer 0) & --- & --- & 16-bit \\
Embeddings / LM head & 16-bit & 16-bit & 16-bit \\
\bottomrule
\end{tabular}}
\caption{Quantization scope, identical for the MxMoE, GEMQ, and \textsc{Q-Strata} rows of \Cref{tab:quant}. Only expert-FFN linears are mixed-precision quantized.}
\label{tab:scope}
\end{table}

\paragraph{Quantization scope.}
\Cref{tab:scope} lists which components are quantized.
The scope is identical for the three mixed-precision methods of \Cref{tab:quant}.
Only the expert-FFN linears of the MoE blocks enter the mixed-precision allocation, and every other linear layer, including the routing gates and the attention projections, stays at 16 bits.

\paragraph{Router fine-tuning.}
For \Cref{tab:gemq} we run GEMQ's official router fine-tuning recipe, training all routing gates for 1 epoch with learning rate $10^{-4}$ on the same calibration data, applied identically to GEMQ's and \textsc{Q-Strata}'s allocations.

\paragraph{Calibration.}
For the experiments on MoE models \Cref{tab:quant,tab:ablation}, all calibration data is drawn from WikiText2 train at sequence length $4096$ following the calibration set preparation protocol of \citet{mxmoe}.
The final GPTQ quantization uses $128$ sequences.
The mixed-precision search uses fewer, $64$ sequences for the inner stage that builds the block-output distortions and $32$ sequences for the outer stage that measures $\mathcal{L}$.
Similarly, for MxMoE, we use $64$ sequences for the inner stage that builds the block-output distortions and quantize the resulting assignment using $128$ sequences.
For GEMQ, we utilize $64$ sequences to obtain a gradient based additive proxy using their implementation and quantize the resulting assignment using $128$ sequences.

\paragraph{Evaluation.}
We report WikiText2 perplexity at context length 4096 and the average of six zero-shot accuracies, PIQA, BoolQ, WinoGrande, ARC-easy, ARC-challenge, and HellaSwag, computed with lm-eval-harness v0.4.5~\citep{eval-harness}. For the JSD objective, we compute JSD between the full precision model and the quantized model with top-$1000$ token logits in MoE experiments for memory efficiency.

\paragraph{Target budgets.}
\Cref{tab:quant} reports target average bitwidths $2.25$, $2.00$, and $1.75$, and \Cref{tab:ablation} reports $2.75$, $2.25$, and $1.75$.

\section{Outer-stage baselines}
\label{app:baselines}
The baselines of \Cref{tab:ablation} share the inner caches $\{\mathcal{S}_l\}$ and differ only in how they pick a budget $\beta_l\in\mathcal{B}$ per block.

\paragraph{Uniform.}
Allocate the same bitwidth for all MoE blocks.

\paragraph{One-shot ILP.}
This baseline modularizes $\mathcal{L}$ once.
From the top assignment $\boldsymbol\beta^{\mathrm{top}}=(\beta^{(K)},\dots,\beta^{(K)})$, for each block $l$ and lower level $\beta^{(k)}$ it measures the marginal $m_{l,k}=\mathcal{L}(x')-\mathcal{L}(x(\boldsymbol\beta^{\mathrm{top}}))$, where $x'$ lowers only block $l$ to $\beta^{(k)}$.
Treating $\sum_l m_{l,k_l}$ as a separable surrogate for $\mathcal{L}$, it selects one level per block by an MCKP under the budget constraint, solved exactly with an ILP.
The surrogate needs only $(K-1)L+1$ evaluations of $\mathcal{L}$, all at the top model, but it assumes the per-block reductions do not interact, an assumption that breaks as several blocks are compressed together.
This is the search used by \citet{qpalette}.

\paragraph{Ascending greedy.}
This is \Cref{alg:outer} run in the opposite direction.
It starts from the cheapest corner $(\beta^{(1)},\dots,\beta^{(1)})$ and repeatedly raises the block whose one-level increase lowers $\mathcal{L}$ the most, until the average bitwidth reaches $\tau$.
It uses the same lazy heap and staleness check as our descent, and differs only in its starting corner and the direction of each move.

\paragraph{Direct cell-level ILP (\Cref{tab:cellilp}).}
This pruning-free baseline measures the change in $\mathcal{L}$ when each (expert, linear layer, bitwidth) cell alone is lowered from the top assignment, and solves the global assignment over raw cells as a single MCKP with an ILP.
The one-shot ILP variant of \Cref{tab:cellilp} applies the same one-shot measurement and ILP on top of our inner-stage caches, so comparing the two isolates the effect of the frontier pruning and the block decomposition.

\section{Additional results}
\label{app:results}
\subsection{Fidelity of the calibration objective}
\label{app:corr}
\begin{table}[t]
\centering
\small
\begin{tabular}{l cc}
\toprule
Score & $\rho$ w/o rot. & $\rho$ w/ rot.\\
\midrule
Calibration JSD & 0.98 & 0.99 \\
GEMQ additive proxy & 0.42 & 0.28 \\
\bottomrule
\end{tabular}
\caption{Spearman rank correlation between each search score and WikiText2 test perplexity over 32 random same-budget allocations (DeepSeek-V2-Lite, 2.0 average bits, per-expert bit maps), each deployed with the final GPTQ pass and evaluated end to end. GEMQ's proxy is computed from its own gradient-weighted statistics on the corresponding (unrotated or rotated) model.}
\label{tab:corr}
\end{table}

To verify that the calibration-set surrogate ranks allocations the way the final metric does, we draw 32 random allocations, all at exactly $2.0$ average bits, on DeepSeek-V2-Lite at GEMQ's native granularity of per-expert bit maps.
% Every allocation is deployed with the corresponding GPTQ quantized model and evaluated on the WikiText2 test dataset.
Every allocation is deployed with the final GPTQ pass and evaluated on the WikiText2 test set.
We then rank the identical allocation set by our calibration JSD and by GEMQ's own search objective, computed from its gradient-weighted per-expert statistics for the same model and calibration data.
In the rotated setting, GEMQ's statistics are recomputed on the rotated model.
The model-level JSD ranks same-budget allocations nearly perfectly by their final test perplexity, with $\rho=0.98$ without rotation and $0.99$ with rotation, while GEMQ's additive proxy correlates far worse on the identical allocations (\Cref{tab:corr}).
% The JSD-searched allocations also attain the best MMLU among the compared quantized methods on all three models at $2.0$ bits (\Cref{app:mmlu}), alongside the zero-shot average gains in \Cref{tab:quant}.

\subsection{MMLU and GSM8K at 2.0 bits}
\label{app:mmlu}
\begin{table}[t]
\centering
\small
\setlength{\tabcolsep}{3pt}
\begin{tabular}{ll cc}
\toprule
Model & Method & MMLU\,($\uparrow$) & GSM8K\,($\uparrow$)\\
\midrule
\multirow{4}{*}{Mixtral}
 & Uniform GPTQ & 24.8 (0.4) & 0.1 (0.1) \\
 & MxMoE & 38.2 (0.4) & 0.8 (0.2) \\
 & \textsc{Q-Strata} & \textbf{42.7} (0.4) & \textbf{3.7} (0.5) \\
 & BF16 & 70.3 (0.4) & 63.5 (1.3) \\
\midrule
\multirow{4}{*}{Qwen1.5-MoE}
 & Uniform GPTQ & 25.6 (0.4) & 0.0 (0.0) \\
 & MxMoE & 34.4 (0.4) & 4.1 (0.5) \\
 & \textsc{Q-Strata} & \textbf{41.6} (0.4) & \textbf{6.0} (0.7) \\
 & BF16 & 61.1 (0.4) & 16.5 (1.0) \\
\midrule
\multirow{4}{*}{DSv2-Lite}
 & Uniform GPTQ & 25.8 (0.4) & 0.0 (0.0) \\
 & MxMoE & 35.8 (0.4) & \textbf{8.4} (0.8) \\
 & \textsc{Q-Strata} & \textbf{41.5} (0.4) & \textbf{8.1} (0.8) \\
 & BF16 & 58.1 (0.4) & 36.5 (1.3) \\
\bottomrule
\end{tabular}
\caption{MMLU / GSM8K (5-shot, \%) at $2.0$ average bits. Standard errors in parentheses. Bold marks the best quantized result and any result within one standard error of it.}
\label{tab:mmlu}
\end{table}
Our allocation deliberately gives some experts very few bits, which raises the question of whether that silently costs capabilities that perplexity misses.
The direct baseline is uniform GPTQ, which spends the identical total budget uniformly across all experts.
We evaluate MMLU and GSM8K, both 5-shot, at $2.0$ bits (\Cref{tab:mmlu}).
The non-uniform allocation preserves these capabilities better, not worse.
Against uniform GPTQ at the same budget, \textsc{Q-Strata} gains $15.7$ to $17.9$ MMLU points on all three models, where uniform GPTQ stays at chance level, and lifts GSM8K from at most $0.1$ to between $3.7$ and $8.1$.
It also attains the best MMLU among the compared quantized methods on all three models, and the best GSM8K among them on two of the three, with DeepSeek-V2-Lite statistically tied with MxMoE ($8.1$ vs $8.4$, within one standard error).

\subsection{Routing distributions of the calibration data}
\label{app:routing}
\begin{table}[t]
\centering
\small
\begin{tabular}{l cc}
\toprule
Model & \makecell{calib vs\\wiki2-test} & \makecell{calib vs uniform\\(reference)}\\
\midrule
Mixtral & 0.009 & 0.042 \\
Qwen1.5-MoE & 0.026 & 0.074 \\
DSv2-Lite & 0.032 & 0.120 \\
\bottomrule
\end{tabular}
\caption{Total-variation distance between expert-routing distributions, per-layer mean ($0$ = identical). The distance from the calibration distribution to the uniform distribution is given as a scale reference.}
\label{tab:routing}
\end{table}

Existing allocation methods including \textsc{Q-Strata} tend to give fewer bits to experts that are routed less frequently on the calibration data, so we check whether the calibration stream is representative of the routing on the test data.
We trace the per-(layer, expert) routing frequencies of the full-precision models and compare the calibration stream against the WikiText2 test set with the total-variation (TV) distance, the fraction of routing mass placed differently (\Cref{tab:routing}).
The calibration routing lies roughly 3 to 5 times closer to the test routing than to the uniform reference.

\subsection{Stability across calibration subsets}
\label{app:seeds}
\begin{table}[t]
\centering
\small
\begin{tabular}{l ccc}
\toprule
Bits & Trial 1 (paper) & Trial 2 & Trial 3\\
\midrule
2.25 & 7.98  & 7.97  & 7.96 \\
2.00 & 8.97  & 9.00  & 8.93 \\
1.75 & 11.84 & 11.91 & 11.83 \\
\bottomrule
\end{tabular}
\caption{The full search pipeline re-run with different calibration subsets (Qwen1.5-MoE, WikiText2 perplexity). Each trial draws the calibration data with a different random seed, and Trial 1 is the paper's run.}
\label{tab:seeds}
\end{table}
As a direct end-to-end check of calibration sensitivity, we re-run the entire \textsc{Q-Strata} search on Qwen1.5-MoE with two fresh calibration sets drawn with different sampling seeds (\Cref{tab:seeds}).
The searched allocations reproduce the paper's perplexity closely, with the three seeds agreeing within $0.08$ perplexity at every budget, so the search outcome on this model is not sensitive to the particular calibration subset.

\subsection{Rotation at extreme bitwidths}
\label{app:rotation}
\begin{table}[t]
\centering
\resizebox{\columnwidth}{!}{
\begin{tabular}{l cc cc cc}
\toprule
& \multicolumn{2}{c}{Mixtral} & \multicolumn{2}{c}{Qwen1.5-MoE} & \multicolumn{2}{c}{DSv2-Lite} \\
\cmidrule(lr){2-3} \cmidrule(lr){4-5} \cmidrule(lr){6-7}
Uniform quantizer & w/o & w/ & w/o & w/ & w/o & w/ \\
\midrule
w4 g128 (4.25 bits) & 3.94 & 3.94 & 6.88 & 6.87 & 5.98 & 5.97 \\
w3 g128 (3.25 bits) & 4.17 & 4.17 & 7.20 & 7.18 & 6.20 & 6.20 \\
w2 g128 (2.25 bits) & 5.73 & 5.73 & 11.24 & 10.41 & 8.70 & 8.59 \\
% w2 per-channel (2.0 bits)$^{\ast}$ & 15.64 & 11.37 & 26.51 & 18.64 & 17.30 & 13.79 \\
w1 g128 (1.25 bits) & 152.5 & 2024.4 & 124495 & 64229 & 3647 & 4262 \\
\bottomrule
\end{tabular}}
\caption{Uniform GPTQ, WikiText2 perplexity, without and with rotation. 
% $^{\ast}$Reference only, not in our quantizer set.
}
\label{tab:unifrot}
\end{table}
To isolate the rotation pipeline from any allocation search, we quantize all three models with uniform GPTQ, with and without the offline random Hadamard rotation of MxMoE's released implementation (\Cref{tab:unifrot}).
Among the group-128 asymmetric quantizers that constitute our quantizer set, rotation consistently matches or improves the rotation-free results at 2, 3, and 4 bits, in line with the common expectation.
% The gain is largest for per-channel quantization, where outliers dominate the quantization scale.
At the extreme 1-bit level, however, the effect of rotation becomes model-dependent, roughly halving the perplexity on Qwen1.5-MoE (124495 to 64229) while severely hurting Mixtral (152.5 to 2024.4).

This model-dependent 1-bit behavior explains \Cref{tab:gemq}.
At the $1.75$- and $2.0$-bit budgets, the searched allocations assign the 1-bit quantizer to 25 to 59\% of the expert linears, forced by the budget itself.
The cheapest quantizer in the shared set costs $1.25$ effective bits and the next-cheapest $2.25$, so an average budget of $1.75$ bits requires at least half of the expert linears at 1 bit.
The rotated \textsc{Q-Strata} results track the 1-bit row of \Cref{tab:unifrot} model by model, with rotation-free winning on Mixtral ($10.22$ vs $12.14$ at $1.75$ bits), the settings tying on DeepSeek-V2-Lite, and rotation winning on Qwen1.5-MoE.

\subsection{Scalability and search cost}
\label{app:scale}
\begin{table}[t]
\centering
\small
\begin{tabular}{ll r}
\toprule
Bits & Method & Wiki2\,($\downarrow$)\\
\midrule
\multirow{3}{*}{2.25}
  & Uniform GPTQ & 12.05 \\
  & MxMoE & 8.73 \\
  & \textsc{Q-Strata} & 8.56 \\
\midrule
\multirow{3}{*}{2.00}
  & Uniform GPTQ$^{\dagger}$ & 14.97 \\
  & MxMoE & 9.25 \\
  & \textsc{Q-Strata} & 8.97 \\
\midrule
\multirow{2}{*}{1.75}
  & MxMoE & 10.48 \\
  & \textsc{Q-Strata} & 10.23 \\
\bottomrule
\end{tabular}
\caption{Qwen3-30B-A3B results, WikiText2 perplexity, under the rotated setting of \Cref{sec:setup}. $^{\dagger}$As in \Cref{tab:quant}.}
\label{tab:qwen3}
\end{table}
\begin{table}[t]
\centering
\small
\begin{tabular}{l ccc}
\toprule
Model & Stage 1 & Stage 2 & Total\\
\midrule
Mixtral-8$\times$7B & 2.5 & 20.5 & 23.0 \\
Qwen1.5-MoE & 12.3 & 11.2 & 23.5 \\
DeepSeek-V2-Lite & 18.0 & 17.5 & 35.5 \\
Qwen3-30B-A3B & 14.8 & 73.9 & 88.7 \\
\bottomrule
\end{tabular}
\caption{Search cost of \textsc{Q-Strata} in GPU-hours, measured on H100. One Stage-2 descent sweeps the entire budget range and emits the allocation at every target bitwidth, so each row is the total cost for all budgets of that model.}
\label{tab:cost}
\end{table}
On Qwen3-30B-A3B, whose 48 blocks of 128 experts hold $18{,}432$ expert linear layers, the outer stage's gains persist, as \textsc{Q-Strata} achieves lower perplexity than the inner-only MxMoE baseline at all three budgets (\Cref{tab:qwen3}).

\Cref{tab:cost} reports the search cost in GPU-hours.
Note that the search is a one-time offline cost, amortized over the deployment lifetime during which the memory and latency benefits of the deployed model accrue (\Cref{app:deploy}).
Producing the deployed model afterwards is a single GPTQ pass, taking less than half an hour on one H100, and the final evaluation is paid identically by every compared method.
The search cost scales linearly with the calibration-set size.

\subsection{Deployment footprint and throughput}
\label{app:deploy}
\begin{table}[t]
\centering
\small
\setlength{\tabcolsep}{4pt}
\begin{tabular}{ll cc}
\toprule
Bits & Method & Model size\,($\downarrow$) & Decode tok/s\,($\uparrow$)\\
\midrule
16 & BF16 & 31.4 GB & 117.9 \\
\midrule
\multirow{2}{*}{2.25}  & GEMQ & 6.0 GB & 173.7 \\
& \textsc{Q-Strata} & 6.0 GB & 172.1 \\

\midrule
\multirow{2}{*}{2.00} & GEMQ & 5.5 GB & 173.9 \\
& \textsc{Q-Strata} & 5.5 GB & 172.0 \\

\midrule
\multirow{2}{*}{1.75}  & GEMQ & 5.0 GB & 172.9 \\
& \textsc{Q-Strata} & 5.0 GB & 171.4 \\

\bottomrule
\end{tabular}
\caption{Model size and decode throughput of the \Cref{tab:quant} allocations of \textsc{Q-Strata} and GEMQ (shared) on DeepSeek-V2-Lite, served with GemLite kernels on one H100 at batch size 1.}
\label{tab:deploy}
\end{table}
Expert weights compress by $7.1\times$ ($16/2.25$) to $9.1\times$ ($16/1.75$) across the four models.
We benchmark decode with GemLite kernels, 1/2/3/4-bit dequantization plus GEMM, which natively serve our quantizer set (\Cref{tab:deploy}).
The quantized models reduce the model size by $5.2$ to $6.3\times$ and speed up decoding by about $1.46\times$ over the BF16 reference.
At a fixed budget the decode speed differs by within about 1\% between the two allocations, so in these measurements the allocation affects quality rather than serving speed.
We leave further optimization of the serving stack, such as CUDA kernels specialized for heterogeneous bitwidths and a kernel-aware formulation of the mixed-precision search, to future work.

\section{Budgeted descent for dense models}
\label{app:dense}

% For a dense model the choice groups are individual layers, which differ in parameter count, so the equal-budget reduction of the main text no longer holds.
% Writing $w_i$ for the parameter count of layer $i$, the budget is the size-weighted mean $\bar b=\sum_i w_i b_i\,/\,\sum_i w_i$, and lowering layer $i$ by one level saves a budget proportional to $w_i$.
% Moves therefore differ in cost, and the descent ranks them by the change in $\mathcal{L}$ per unit budget saved, $\Delta\mathcal{L}_i/(w_i\Delta)$  \citep{khuller1999}.
% Concretely, \Cref{alg:outer} keeps the min-heap keyed by this normalized value rather than by $\Delta\mathcal{L}$ alone, and the stopping test compares the size-weighted mean against $\tau$.
% The lazy refresh and the staleness check are unchanged.

The outer descent needs one adjustment on dense models.
In the MoE setting every block holds the same number of parameters, so every one-level move frees the same amount of budget, and \Cref{alg:outer} ranks moves by the loss increase alone.
On a dense model the choice groups are individual layers, and layers differ in parameter count.
Writing $w_i$ for the parameter count of layer $i$, the average bitwidth becomes the size-weighted mean $\bar b=\sum_i w_i b_i\,/\,\sum_i w_i$, and lowering layer $i$ by one level frees a budget proportional to $w_i$.
A move on a large layer therefore saves more budget than one on a small layer, and the descent ranks moves by the loss increase per unit of budget freed, $\Delta\mathcal{L}_i/(w_i\Delta)$, where $\Delta$ is the spacing of the budget grid~\citep{khuller1999}.
\Cref{alg:outer} changes in two places only.
The min-heap is keyed by this ratio instead of $\Delta\mathcal{L}_i$ alone, and the stopping test compares $\bar b$ with $\tau$.
The lazy refresh and the staleness check are unchanged.

\paragraph{Dense protocol.}
We use HQQ quantizers at $2$, $3$, and $4$ bits with group size $128$ and asymmetric as the quantizer set.
We calibrate on WikiText2 train, $2$K tokens for \Cref{tab:dense-search} and $8$K tokens for \Cref{fig:hqq_l27b_search_cmpr}, and report perplexity on WikiText2 test set at context length $2048$.
All methods optimize the same JSD objective on the same calibration data and differ only in how many times they evaluate it, as listed per method.

\paragraph{Query counts.}
Every method returns a full set of budgets from one run, so each count in \Cref{tab:dense-search} is the total for that single search rather than a per-budget figure.
For eager greedy and lazy greedy this is one descent from $4.25$ bits to the $2.5$-bit target, which passes through every intermediate budget and yields the whole frontier at once.
One-shot ILP fixes its surrogate with $(K-1)L+1$ measurements and re-solves the ILP per budget at no further query cost.

\end{document}